\documentclass[letterpaper, 10 pt, journal, twoside]{ieeetran}
\IEEEoverridecommandlockouts                              % This command is only needed if 
\usepackage{stfloats}
\usepackage[utf8]{inputenc}
\usepackage{graphicx} 
\usepackage{color}
\usepackage{comment}
\usepackage{amssymb} %% The amssymb package provides various useful mathematical symbols
\usepackage{amsmath}
\usepackage{multirow}
\usepackage{float}
\usepackage{bm}
\usepackage{cancel}
\usepackage{subfigure}
\usepackage[dvipsnames]{xcolor}
\usepackage{cite}
\usepackage{cuted}
\usepackage{fancyhdr}
\usepackage{etoolbox}
\usepackage{hyperref}

\usepackage[top=2.01cm, bottom=1.52cm, left=1.69cm, right=1.69cm]{geometry}
\usepackage[absolute,overlay]{textpos} % For copyright

\usepackage{algorithm}
\usepackage[noend]{algpseudocode}

\title{\LARGE \bf
A novel concept for Titan robotic exploration based on soft morphing aerial robots}

\author{F. Ruiz$^{1}$, H. Yang$^{1}$, B.C. Arrue$^{1}$ and A. Ollero$^{1}$ \\
 $^{1}$ GRVC Robotics Lab, University of Seville, Spain. \\
IAC-23 - A.3.5 - Solar System Exploration Including Ocean Worlds}

\begin{document}

\maketitle

% \thispagestyle{empty}
% \pagestyle{empty}

%%%%%%%%%%%%%%%%%%%%%%%%%%%%%%%%%%%%%%%%%%%%%%%%%%%%%%%%%%%%%%%%%%%%%%%%%%%%%%%%

\vspace*{-5mm}

\begin{strip}

\begin{abstract}

This work introduces a novel approach for Titan exploration based on soft morphing aerial robots leveraging the use of flexible adaptive materials. The controlled deformation of the multirotor arms, actuated by a combination of a pneumatic system and a tendon mechanism, provides the explorer robot with the ability to perform full-body perching and land on rocky, irregular, or uneven terrains, thus unlocking new exploration horizons. In addition, after landing, they can be used for efficient sampling as tendon-driven continuum manipulators, with the pneumatic system drawing in the samples. The proposed arms enable the drone to cover long distances in Titan's atmosphere efficiently, by directing rotor thrust without rotating the body, reducing the aerodynamic drag. Given that the exploration concept is envisioned as a rotorcraft planetary lander, the robot's folding features enable over a 30$\%$ reduction in the hypersonic aeroshell's diameter. Building on this folding capability, the arms can morph partially in flight to navigate tight spaces. As for propulsion, the rotor design, justified through CFD simulations, utilizes a ducted fan configuration tailored for Titan's high Reynolds numbers. The rotors are integrated within the robot's deformable materials, facilitating smooth interactions with the environment. The research spotlights exploration simulations in the Gazebo environment, focusing on the Sotra-Patera cryovolcano region, a location with potential to clarify Titan's unique methane cycle and its Earth-like features. This work addresses one of the primary challenges of the concept by testing the behavior of small-scale deformable arms under conditions mimicking those of Titan. Groundbreaking experiments with liquid nitrogen at cryogenic temperatures were conducted on various materials, with Teflon (PTFE) at low infill rates (15-30$\%$) emerging as a particularly promising option due to its thermal stability. 

%The results demonstrate encouraging prospects in terms of power consumption and exploration capabilities when compared to NASA's planned Dragonfly mission set for 2027.

%The folding capability of the robot also relaxes hypersonic aeroshell design constraints. The rotors are embedded in
%the deformable material to guarantee a smooth interaction with the environment, which is leveraged to access caverns or confined spaces.

%\newline
%\newline
%%\textit{Index Terms:} Space Exploration, Soft Robotics, Mechanical Design, Aeroelasticity

\end{abstract}

\end{strip}

%%%%%%%%%%%%%%%%%%%%%%%%%%%%%%%%%%%%%%%%%%%%%%%%%%%%%%%%%%%%%%%%%%%%%%%%%%%%%%%%
\section{INTRODUCTION}

\IEEEPARstart{S}{pace} exploration has consistently been at the forefront of scientific endeavors, deepening the understanding of the composition, atmospheres, geological phenomena, and specific characteristics of various planetary bodies within the solar system. From inner planets like Mercury, Venus, and Mars, to distant gas giants and their complex moon systems, understanding these celestial entities has been of paramount importance to global space agencies. This quest for knowledge has led to the deployment of diverse exploration methodologies, ranging from Earth-based telescopic observations \cite{Ostro19931235, Telescope1} to complex missions involving orbiters \cite{biesbroek2000ways, Guillot199972}, landers \cite{Gromov199851, Morris20001757}, and rovers \cite{KUBOTA2003447, Curiosity}.

Recently, the potential of aerial robotic systems, particularly Unmanned Aerial Vehicles (UAVs), has emerged as a significant focus in space exploration. With their capability to provide detailed, high-resolution data from close proximity to planetary surfaces, drones are now being considered for various missions \cite{barret2000aerobots, Elfes2003147}. The conceptualization of these drones spans from fixed to flapping and rotary wing designs, each tailored to operate in distinct planetary atmospheres such as those on Mars, Venus, and Titan \cite{HASSANALIAN201861}. While these UAVs hold tremendous promise, they come with inherent challenges in terms of deployment, operational dynamics, and data transmission.

Saturn's moon Titan is one of the celestial bodies that most closely resembles Earth, possibly the most similar in the entire solar system. Exceeding the size of Mercury, Titan is shrouded in a thick nitrogen-dominated atmosphere interspersed with organic smog. This veil makes direct observations of its surface challenging, yet intriguing. Distanced from the Sun, Titan's frigid environment allows methane to play the role water does on Earth: it acts as a greenhouse gas, forms clouds, and settles in lakes and seas \cite{Titan1, Titan2}.

\begin{figure}
\includegraphics[width=0.499\textwidth,scale=0.25]{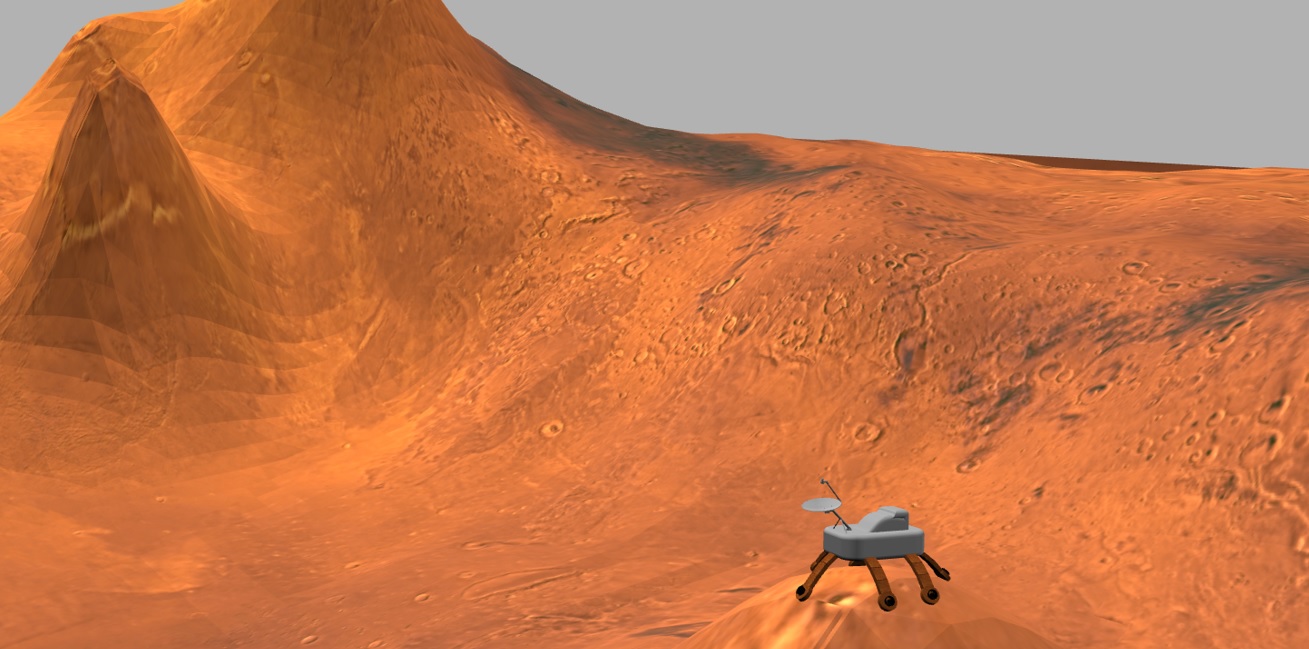}
\caption{The novel soft morphing aerial robot exploring Titan's Sotra Patera region, landing on a rocky and irregular surface using its flexible arms (Gazebo environment). Landing video link: \url{https://youtu.be/nf-6Mm_qR-Y}. Forward flight video link: \url{https://youtu.be/seYnK68ZQo4}.}
\label{figure1}
\vspace{-2mm}
\end{figure}

The topography of Titan is a blend of impact craters, expansive linear dunes shaped by winds, methane-carved river channels, and potential sites of cryovolcanism. From an astrobiological perspective, Titan stands out as an "ocean world" rich in carbon and nitrogen, unlike Europa's water and sulfur abundance. Especially based on the data from the Huygens probe \cite{TITAN3}, most of Titan's features like its dense atmosphere, which is four times denser than Earth's sea-level air, and a surface gravity of 1.35 $m/s^2$, were discovered. Despite receiving light levels about 1000 times fainter than Earth, primarily in red and near-infrared wavelengths, Titan remains a subject of priority scientific interest.

Within the realm of autonomous vehicles tailored for extraterrestrial exploration, Titan presents a set of unique challenges and opportunities. Historically, the designs of such vehicles have been primarily oriented to cater to specific terrains or domains. For instance, landers and rovers, as referenced in \cite{TITAN3}, are typically designed to traverse solid, regular surfaces. Underwater vehicles aim to study the oceans, as mentioned in \cite{Stofan2010}. Lastly, atmospheric probes, including both balloons \cite{Balloons} and drones such as those in \cite{ShapeShifter, Dragonfly, AVIATR}, are specifically engineered to study and navigate Titan's atmosphere.

\begin{figure*}[t]
    \centering
    \includegraphics[width=0.94\linewidth]{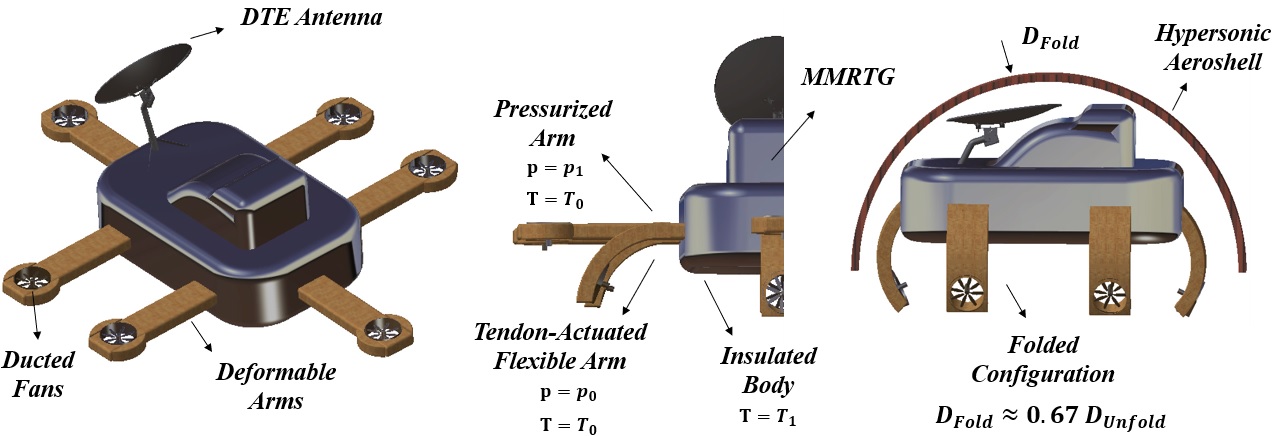}
     \caption{Conceptual aerial robot design for Titan robotic exploration. The main body is maintained at mild temperatures ($T_1$) while the arms are at ambient temperature $T_0=-180$ degrees. During flight, the pressure inside the arms ($P_1$) is controlled and is set above $P_0$ to adjust the flexibility of the arms, which can be actuated by a tendon system. The mission concept is that of a rotorcraft planetary lander, so the robot's folding capability reduces hypersonic aeroshell constraints. }
\label{figure1}
\vspace{-1mm}
\end{figure*}

The most advanced concept to date corresponds to NASA's Dragonfly mission \cite{Dragonfly}, scheduled for 2027. This is a conventional multirotor vehicle capable of carrying all necessary instruments to take surface samples and transmit data to Earth using an antenna. However, there are other, riskier drone proposals that might be more efficient and adaptable, such as the Shapeshifter \cite{ShapeShifter}. This innovative robotic system comprises several "Cobots" or robotic units. These units possess the unique capability to magnetically connect, forming a range of configurations, from rolling vehicles and flying arrays to potential underwater structures. To support its functionality, the Shapeshifter features a landing module, responsible for both recharging and communication tasks.

However, despite these advancements, to the authors' best knowledge, the literature still lacks a comprehensive vehicle design that can seamlessly land on Titan's irregular surfaces and adapt accordingly. Furthermore, when exploring the hard-to-access places or caverns of Titan, drones with exposed rotors rotating at high speeds pose too much risk for such tasks. %Additionally, the remaining challenge lies in achieving all of this while maintaining a low weight and minimizing hypersonic aeroshell requirements.

The use of adaptive flexible materials in the design has the potential to address these challenges. The incorporation of soft materials in drones is a significant and advancing research area \cite{Floreano_origami, MINICORE}, to the extent that a fully flexible UAV has been developed \cite{Ruiz2022b}, ensuring its functionality from both a mechanical \cite{Ruiz2022a} and control standpoint \cite{Ruiz2022c}. Some of the properties of these drones can be particularly relevant for the inherent characteristics of space exploration.

However, this poses certain challenges due to Titan's extreme conditions, especially its low temperatures around -180 degrees. Most flexible materials crystallize at these temperatures, becoming rigid and brittle \cite{TPU1,silicone1,cryogenics1}. This is an ongoing research area of particular interest for space exploration because of the potential that flexible materials might have in these environments \cite{Teflon1,TPU2}.

In this context, this work proposes the use of adaptive flexible materials in UAVs for space exploration for the first time. By merging a pneumatic system with a tendon mechanism, the robot's multirotor arms can adapt their form, enabling unique landing capabilities on uneven terrains and expanding exploration possibilities (see Figure 1), also aiding in the sampling collection process. These arms also enhance flight efficiency by allowing fully-actuated locomotion and reducing aerodynamic drag. Their folding capability reduces the requirements of the planetary lander and facilitates navigation in confined spaces. Another key feature of the concept is the rotor design, optimized for Titan's conditions, which is integrated within the robot's flexible structure, fostering smooth interactions with the environment. 

The methodology followed to validate the concept is based on Gazebo simulations focused on Titan's intriguing Sotra-Patera region, CFD simulations to justify the rotors design (a ducted fan configuration is more efficient on Titan due to its high Reynolds numbers), and finally, an experimental part testing the adaptability of small-scale deformable arms under Titan-like conditions. This involved pioneering experiments with liquid nitrogen, spotlighting PTFE with 15-30$\%$ infill as a promising material due to its thermal resilience.   

In this way, the document is structured as follows: the introduction discusses the state of the art in aerial vehicles for space exploration, specifically for Titan. Section II delves into the concept description, including specific features and designs. Section III presents the experimental analysis results of the behavior of small-scale deformable arms at cryogenic temperatures using liquid nitrogen. Section IV examines the aerodynamics of embedded rotors in Titan's atmosphere using CFD techniques, resulting in the optimal design in terms of diameter and number of blades. In Section V, Gazebo simulations of the landing and morphing capabilities are presented, along with a comparison with other concepts in terms of energy consumption and capabilities. Finally, Section VI presents the conclusions of the work.

\section{Proposed concept}

NASA's Dragonfly mission is scheduled for 2027, with a projected arrival on Titan around 2034. The project was solidified after NASA initiated a competition inviting proposals for potential mission designs and aerial robotic vehicle concepts suitable for Titan exploration. The mission is set to land in Shangri-La, a dark equatorial region on Titan filled with vast dunes and organic material (see Figure 3). This landing site offers an ideal environment to study the prebiotic chemistry processes, with its combination of vast dunes and potential subsurface water-ice reservoirs. It is located near the plain where the Huygens probe landed in 2005. 

\begin{figure}
\includegraphics[width=0.492\textwidth,scale=0.25]{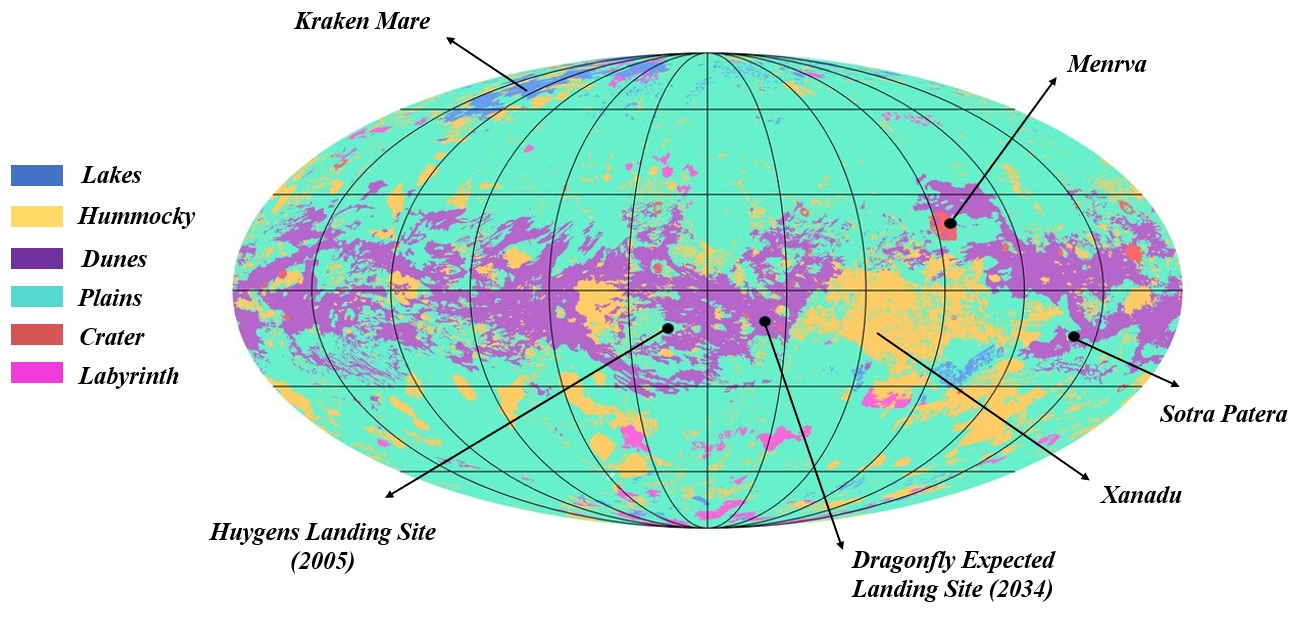}
\caption{Map of Titan based on data from the Cassini mission spanning several years (NASA/JPL-Caltech). The diverse terrains – including lakes, labyrinth areas, hummocky terrains, plains, dunes, and craters – are color-coded for distinction. Key landmarks such as the Huygens and Dragonfly landing sites, along with significant regions like Sotra Patera and Xanadu, are clearly marked.}
\label{figure1}
\vspace{-1mm}
\end{figure}

However, such a design might present inherent landing limitations once it begins its operations in Titan's atmosphere, being primarily restricted to regular plains and dunes. Titan hosts several other regions of great interest, such as the mysterious highlands of Xanadu, known for its potential evidence of liquid water or hydrocarbon-based life, and the cryovolcanic area of Sotra Patera, which could provide insights into Titan's internal geology and potential subsurface ocean. Additionally, the moon features labyrinth terrains, a complex network of valleys and ridges. In all these areas, characterized by craters, labyrinth-like formations, and hummocky terrains (rolling terrains with small elevation changes), finding suitable landing spots can be a significant challenge, and sampling opportunities may be limited.

This work proposes a novel concept whose primary innovation is the inclusion of flexible arms made from adaptive materials. These enhancements aim to expand the robot's exploration potential by equipping it with full-body perching capabilities, allowing it to land on rough terrains, access confined or tight spaces thanks to morphing abilities, increase efficiency, and facilitate effective sampling.

Specifically, by manipulating the deformation of the multirotor arms, primarily through a pneumatic system combined with a tendon-driven mechanism, the explorer robot can be endowed with these capabilities. Furthermore, these features enhance the efficiency of the sampling process by utilizing the robot's soft arms as continuum manipulators and leveraging the pneumatic system to collect samples. To ensure smooth interactions with its environment, rotors are embedded within the flexible material in a ducted fan configuration, more efficient for Titan's dense and low-viscosity atmosphere due to its operation at high Re on Earth, just as the low Re insect-like flapping wings are theoretically more efficient in an atmosphere like that of Mars. This will be discussed in detail in section IV.

Figure 2 presents the conceptual design of an aerial robot intended for Titan exploration. The robot's core body is kept at moderate temperatures ($T_1$) due to a robust insulation, reminiscent of the Huygens probe. This protective layer shields the onboard electronics and primary instruments, encompassing a mass spectrometer, various sensors, cameras, among others. The excess heat from the MMRTG, the principal energy source onboard (as solar energy isn't feasible on Titan), ensures these systems remain in temperate conditions. Conversely, the robot's arms are exposed to Titan's ambient temperature, $T_0 \approx -180$ degrees. During its aerial operations, the internal pressure of the arms ($P_1$) is meticulously managed to be higher than $P_0$, allowing for adjustment of the arms' flexibility, which can be actuated by a tendon mechanism. This pneumatic system is also harnessed for sampling, using the arms as continuum manipulators, controlled by the tendons, to select the exact sampling location.

Conceptually, this robot is visualized as a rotorcraft planetary lander. Its ability to fold is an asset, minimizing hypersonic aeroshell restrictions (the diameter can be reduced up to 33$\%$). Moreover, the mission prototype for Titan requires long-distance atmospheric flights. For this purpose, the proposed arms allow the drone to cover large distances efficiently by directing the rotor thrust without rotating the body, thereby reducing aerodynamic drag. As will be elaborated upon in Section III, the proposed material for these adaptive arms is 3D-printed PTFE, with a sparse infill rate of 15-30$\%$, guaranteeing consistent thermal resilience.

\section{Experimental characterization of small-scale deformable arms}

A significant challenge of the proposed design is how the flexible arms will react in Titan's cold environment, with temperatures close to -180 degrees Celsius. At such low temperatures, many flexible materials might lose their elasticity, turning brittle and increasing their stiffness significantly. In this work, several smaller-scale flexible arms are fabricated using various materials, driven by tendons, to assess their behavior in cold conditions. Specifically, the materials analyzed included TPU at low infill rates, silicone elastomer, and Teflon.

Figure 4 displays the test bench used to assess the behavior of the small-scale deformable arms. This test bench consists of an aluminum rod structure within a container made of the same material, ensuring safety when working with liquid nitrogen. The nitrogen is stored in a double-walled Dewar tank at -196°C. This tank features an intermediate vacuum to eliminate heat transmission via conduction and convection, and an internal reflective surface to minimize radiation. A Type-T thermocouple is used, valued for its precision in tracking temperatures in severe conditions. The setup also incorporates the high-torque Hitec D950TW servo to control the force exerted by the tendons. This figure also showcases the produced small-scale deformable arms: TPU and PTFE are manufactured through FDM, while silicone is shaped by molding. During these tests, cryogenic safety goggles and gloves are worn to ensure safety. 

\begin{figure}
\includegraphics[width=0.492\textwidth,scale=0.25]{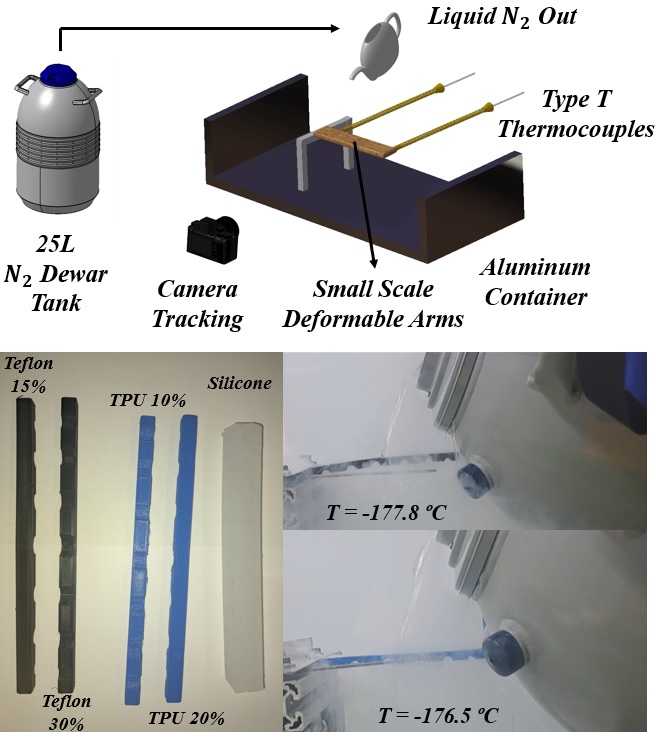}
\caption{Test bench setup and diagram of the process for conducting experiments with deformable arms at cryogenic temperatures using liquid nitrogen (above). The figure below displays the various small-scale deformable arms produced and tested, along with images of their cooling process using liquid nitrogen.}
\label{figure1}
\vspace{-1mm}
\end{figure}

As shown in Figure 5, the evolution of the servomotor force (kgcm) per unit of curvature for these small-scale deformable arms (1/cm) is presented across a broad range of cryogenic temperatures. The results show a clear crystallization of TPU and silicone at around -120 and -96 degrees Celsius, respectively. The arms become much stiffer and harder to deform, unlike the results at room temperature shown in Figure 6. On the other hand, PTFE remains extremely stable against temperature changes, only increasing by 15$\%$ across the entire range studied. This positions it as the ideal candidate, especially at low infill rates, where it becomes somewhat more flexible and with a considerably lower weight.

\section{Rotor aerodynamics}

Ducted fans are a type of propulsion system characterized by a shroud or duct that surrounds a series of shorter and more numerous blades, distinguishing it from traditional propellers. Ducted fans can mitigate the tip losses from the blades; however, while in Earth's atmosphere, they aren't as efficient as propellers for generating static lift, as seen in drones. Even though they can achieve the necessary thrust with smaller blade diameters, lower torque motors, and by compensating with higher RPMs and additional blades, this setup reduces their efficiency compared to standard propeller blades, especially in drone applications. This is mainly due to aerodynamic losses at extremely high rotational speeds and the increase in the number of blades. Nevertheless, ducted fans offer a safety advantage since their design inherently shields the blades.

\begin{figure}
\includegraphics[width=0.492\textwidth,scale=0.25]{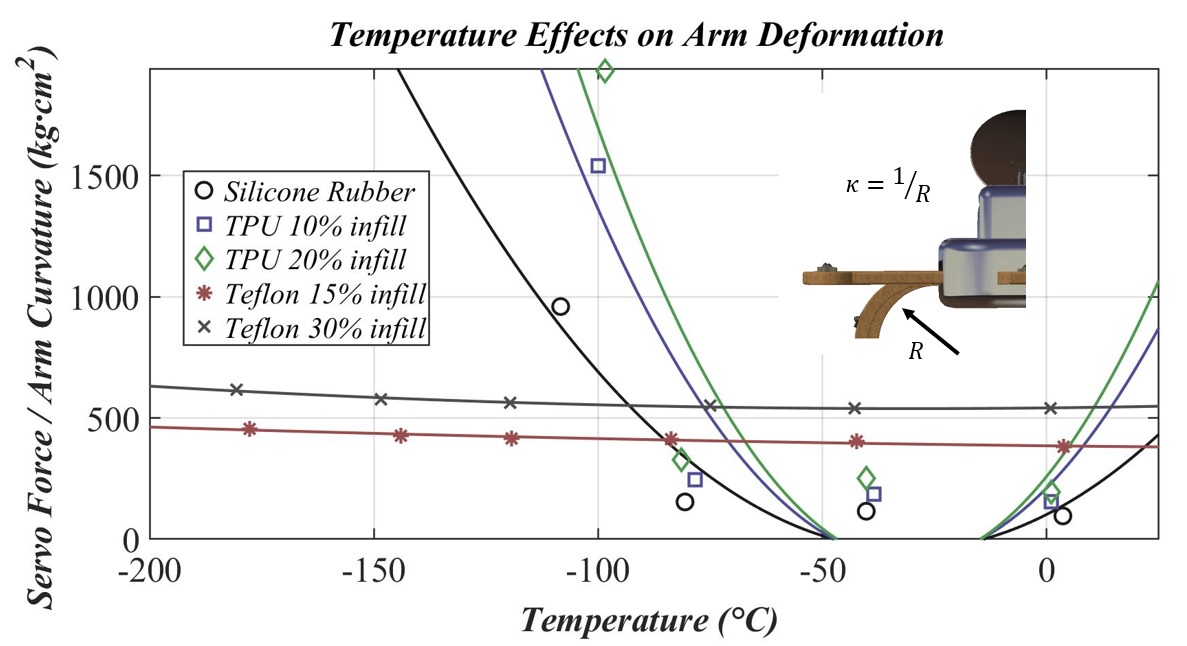}
\caption{Evolution of the servomotor force (kgcm) per unit of curvature  (1/cm) of the small-scale deformable arm over a range of cryogenic temperatures down to -180 degrees, for the proposed materials (TPU at 10$\%$ and 20$\%$ infill, PTFE at 15$\%$ and 30$\%$ infill, and silicone elastomer).}
\label{figure1}
\vspace{-1mm}
\end{figure}

\begin{figure}
\includegraphics[width=0.492\textwidth,scale=0.25]{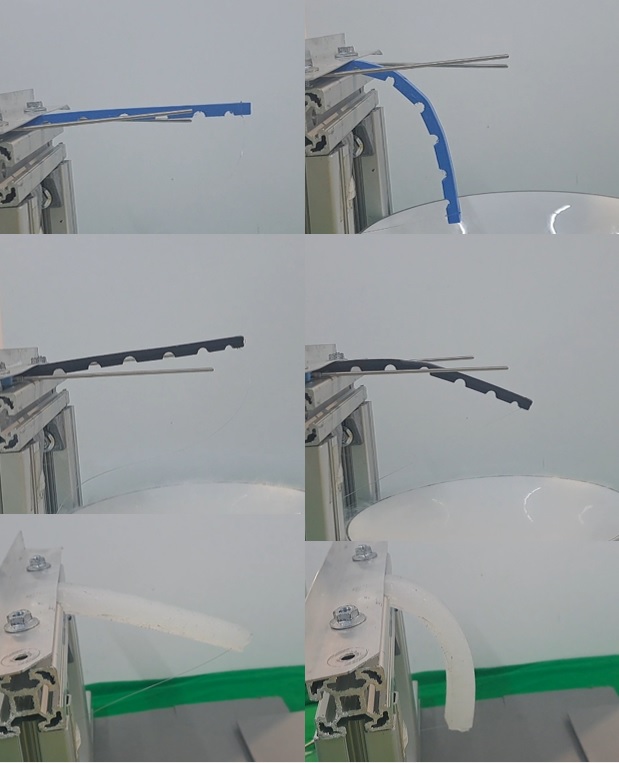}
\caption{Small-scale deformable arms actuated by tendons on the test bench to evaluate the ratio between the servomotor force and deformation curvature at different temperatures. From top to bottom: TPU, teflon, silicone.}
\label{figure1}
\vspace{-1mm}
\end{figure}

\begin{figure}
\includegraphics[width=0.492\textwidth,scale=0.25]{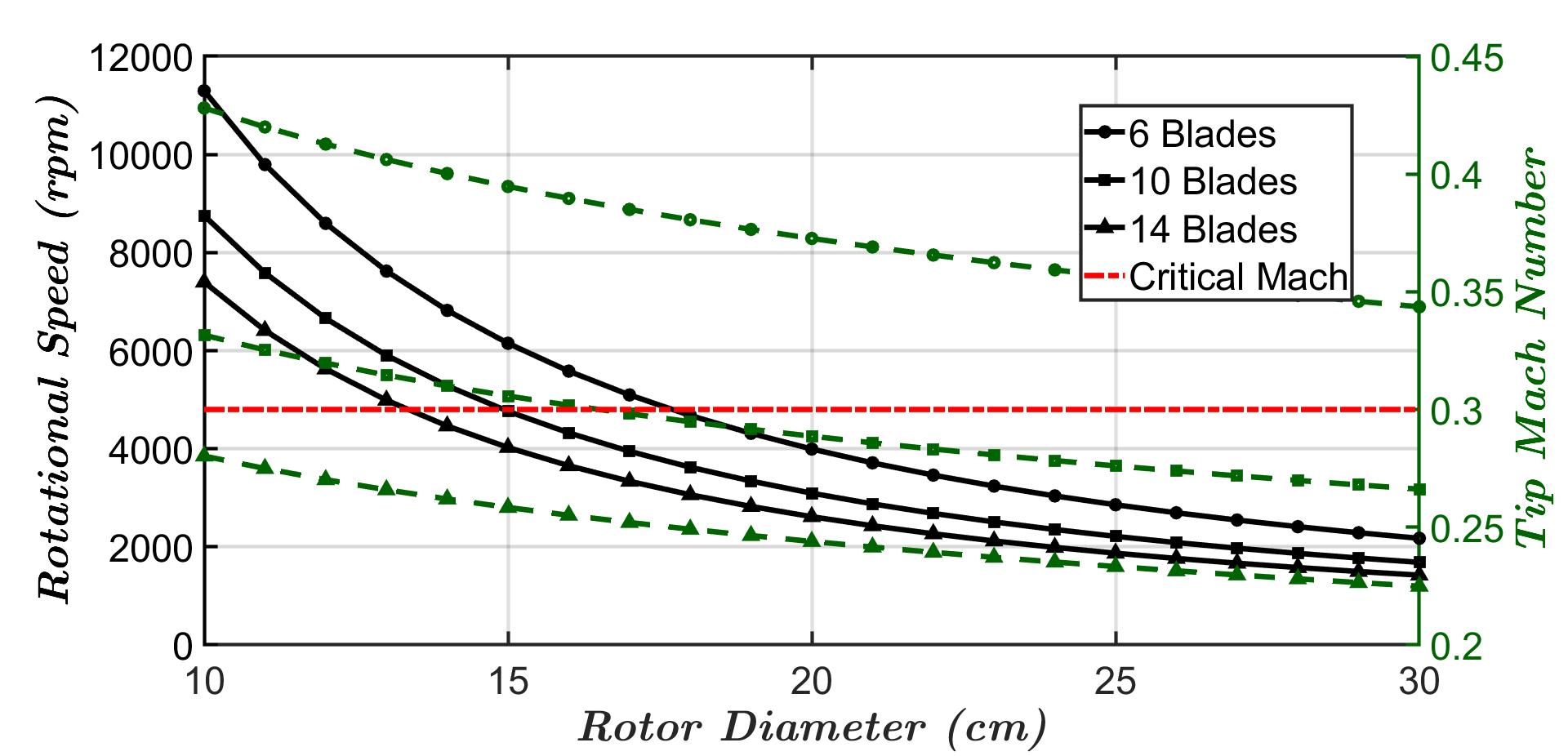}
\caption{Characterization of the ducted fan in Titan's atmosphere for an imposed thrust of 100 N, with the rotor diameter and the number of blades as free variables. The rotational speeds and the Mach number at the tip were derived through interpolation from discrete CFD results (Table I). The critical Mach number is also shown.}
\label{figure1}
\vspace{-2mm}
\end{figure}

In Titan's atmosphere, with a density four times greater and a viscosity three times lower, the Reynolds number 
$Re =\frac{\rho u L }{\mu}$ (where $\rho$ is density, u is rotational velocity, L is characteristic length, and $\mu$ is dynamic viscosity) for a given rotational speed can be up to 12 times higher. This makes the typically high Re operating ducted fans more efficient on Titan, just as the low Re insect-like flapping wings are theoretically more efficient in an atmosphere like that of Mars: ducted fan blades might rotate 12 times slower. Additionally, the torque produced by conventional propellers in Titan's atmosphere would be exceedingly high due to these densities, potentially leading to mechanical issues over time, as well as complicating yaw control due to the lack of sensitivity. The lower torques of ducted fans would help mitigate this issue.

In summary, given that the fixed variable is the total drone weight (420 kg), which is mandated by the instruments of the Dragonfly mission, this translates to a required thrust of 100 N per rotor, for the proposed 6 ducted fan configuration. The free variables include the rotor diameter and the number of blades. Figure 7 displays the results obtained when setting both variables. The curves show the corresponding rotational speed and the Mach number at the tip. For this, nine configurations have been simulated in CFD, corresponding to 3 diameters (15,20 and 25 cm) and 3 numbers of blades (6, 10, and 14), as shown in Table I alongside the associated torques. Polynomials were then fitted to these points to obtain the displayed curves.

To choose the optimal configuration, different sources of inefficiency must be taken into account, especially since, on Titan, the speed of sound is lower, around 194 m/s. This poses constraints on the minimum possible diameter because at very high rotational speeds, the Mach number at the tip increases, leading to aerodynamic losses. The results from Figure 7 suggest that, considering a critical Mach of 0.3, the most suitable configuration corresponds to the diameter of 22 cm with 10 blades, rotating at approximately 3000 rpm with a torque of 4.7 Nm.

\begin{table}[t]
\caption{CFD-Simulated rotor configurations and torque calculations for the soft aerial robot explorer}
\centering
\begin{tabular}{ c  c c c c c  }
\hline
Thrust & Blades &  Diameter & Torque & RPM   \\
\hline
100 N & 6 & 15 cm & 3.7 Nm & 6080 \\
100 N & 10 & 15 cm & 3.9 Nm & 4900   \\
100 N & 14 & 15 cm & 4.05 Nm & 3950  \\
100 N & 6 & 25 cm & 5.6 Nm & 3100   \\
100 N & 10 & 25 cm & 5.9 Nm & 2550  \\
100 N & 14 & 25 cm & 6.3 Nm & 1950   \\
\hline
\end{tabular}
\label{table1}
\vspace{-4mm}
\end{table}

\begin{figure*}
\centering
\includegraphics[width=0.9\textwidth,scale=0.25]{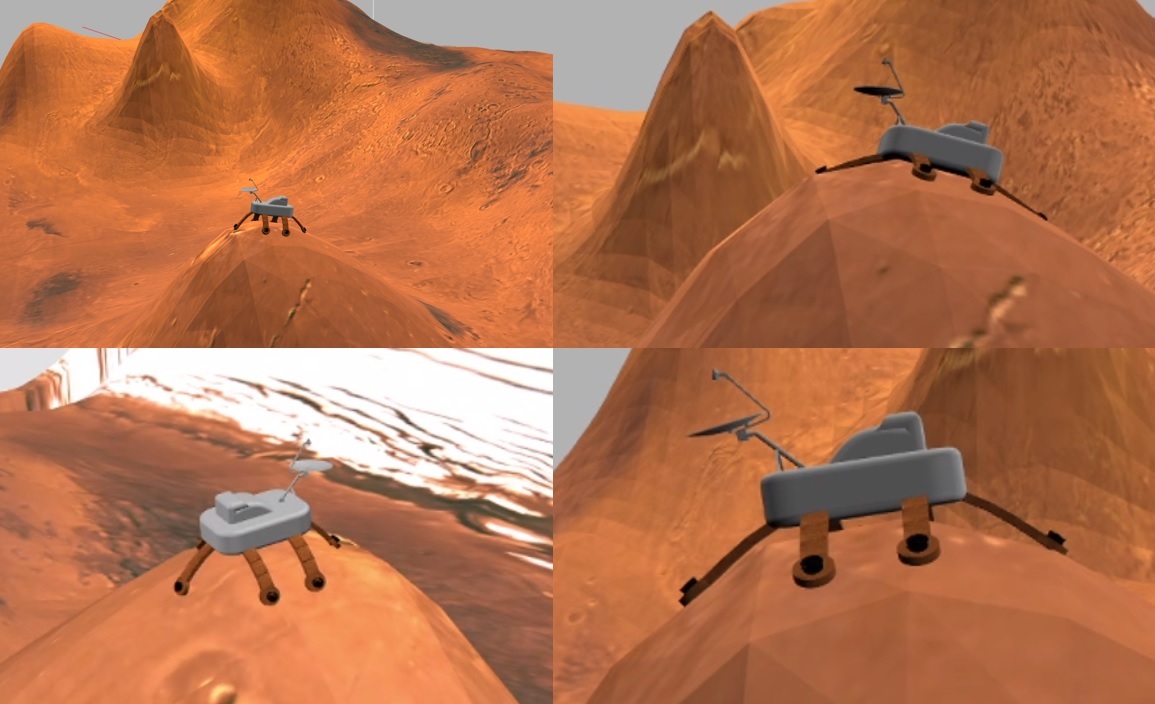}
\caption{Simulations in Gazebo illustrating the soft morphing aerial robot's adaptive landing capabilities across diverse rocky and uneven terrains within Titan's Sotra Patera region.}

\label{figure1}
\vspace{-1mm}
\end{figure*}

\section{Performance and Results}

\subsection{Gazebo Simulations in Titan's Environment}

This section delves into the performance assessment of the proposed soft morphing aerial robot designed for exploration on Titan. Comprehensive simulations were executed in Gazebo, with a keen focus on maneuvers within Titan's Sotra Patera region. The simulations aimed to evaluate the robot's capabilities across multiple scenarios:

\begin{itemize}
    \item \textbf{Landing Dynamics:} Understanding the robot's ability to safely land on a variety of terrains, from rocky outcrops to icy plains, is critical. The morphing feature of the arms plays a significant role in ensuring soft landings, even on irregular surfaces.
    
    \item \textbf{Fully-Actuated Lateral Motion:} Beyond typical aerial maneuvers, the robot's design facilitates efficient side-to-side movements without a need to tilt or roll. 
    
    \item \textbf{Morphing in Flight:} The unique arm design and the embedded rotors can change their configuration during flight. This adaptability enables the robot to alter its shape, optimizing it for different flight modes or to interact with its surroundings more efficiently.
\end{itemize}

To closely replicate Titan's diverse landscape, the Gazebo world was equipped with a variety of terrains, including rocky outcrops, icy patches, and others. These features ensured that the robot was tested under conditions mimicking the complexities it would face on the actual moon.

The unique deformable arms of the robot are one of its defining features. In an attempt to emulate this flexibility, the arms were modeled with multiple joints distributed along their lengths. By applying bending moments to these joints, the soft morphing capabilities of the arms, achieved through tendon mechanisms, were simulated. This arrangement offers an approximation to the infinite degrees of freedom that a real flexible arm would exhibit.

Regarding the simulation environment setup, the PX4 software was employed as the flight controller for the drone. This choice was motivated by PX4's capability to ensure accurate and realistic flight dynamics during the tests. Additionally, the open-source nature of PX4 allows for easy modification of its source code. This adaptability proves especially valuable when introducing specific changes, such as alterations in the mixing matrix for thrust vectoring. As such, it provides a robust and flexible platform to assess the robot's performance in diverse flight conditions.

\subsection{Expanding Exploration Opportunities through Landings on Uneven or Rocky Surfaces}

The flexible drone designed for Titan represents a significant advancement in extraterrestrial exploration tools. Its deformable arms, capable of adapting to the landscape, provide a distinctive advantage: full-body perching. Unlike traditional rigid drones equipped with standard landing gear, which may struggle in challenging terrains such as Titan's labyrinth terrains or hummocky areas, the proposed drone, with its tendon-controlled arms, can securely grip and perch on these surfaces.

Beyond just offering a technical advantage, this capability can fundamentally expand the scope of scientific exploration on Titan. For instance, while the primary simulations focus on the Sotra Patera region – a region replete with geological wonders and indicative of Titan's atmospheric and geological dynamics – the drone's design provides the potential to explore other captivating areas like Xanadu's highlands.

The dynamics of the landing process itself have been meticulously simulated. During the landing phase, the tendons in the drone's arms maintain tension, ensuring a secure grip on the landing surface. However, looking forward to actual implementation, it is envisioned that a locking mechanism would be integrated. This mechanism would allow the drone's arms to maintain the position post-landing, ensuring stability without continuously expending energy to maintain grip tension.

Figure 8 provides a visual representation of the drone's landing simulations within the Sotra Patera region, exemplifying its ability to adeptly handle diverse and challenging terrains that are characteristic of Titan's surface.

\subsection{Leveraging Deformable Arms for Efficient Fully-Actuated Mobility on Titan's Surface}

Fully-actuated flight with deformable arms can significantly enhance flight efficiency. This means that the drone can traverse Titan's atmosphere horizontally without needing to rotate its body to tilt the rotors; only the arms perform this motion, as depicted in Figure 9. This approach minimizes the frontal area (and drag coefficient) during such movements, substantially boosting aerodynamic efficiency (as long as the design is careful to avoid undesired aerodynamic interactions that generate inefficiencies). Given the mission profile for Titan, which necessitates covering long distances (spanning hundreds of kilometers) through its atmosphere, this attribute is particularly crucial.

Hence, this section evaluates the power consumption of the soft morphing aerial robot and compares the fully-actuated case to the rigid one. The methodology followed is the same as described in \cite{energetics_titan}. The power consumed during flight is divided into three primary components: body drag power, induced power, and profile power. Each of these is influenced by various aerodynamic and geometric properties of the drone.

\textbf{Body Drag Power $P_{\text{D}}:$} This is due to the aerodynamic resistance experienced by the drone's body. Mathematically, it's represented as:

\begin{equation}
P_{\text{D}} = 0.5 \rho v^3  S  C_D
\end{equation}

Where:
\begin{itemize}
    \item \(\rho\) is Titan`s density.
    \item \(v\) is the drone's velocity.
    \item \(S\) is the drone's frontal area.
    \item \(C_D\) is the drag coefficient of the drone's body.
\end{itemize}

\textbf{Induced Power $P_{\text{I}}$:} It is the power required to overcome the induced drag generated by the creation of lift. Utilizing the momentum theory for hovering drones $P_{\text{I\_h}}$:

\begin{equation}
P_{\text{I\_h}} = \sqrt{(m g)^3  \rho / 2 / A}
\end{equation}

\begin{figure}
\includegraphics[width=0.492\textwidth,scale=0.25]{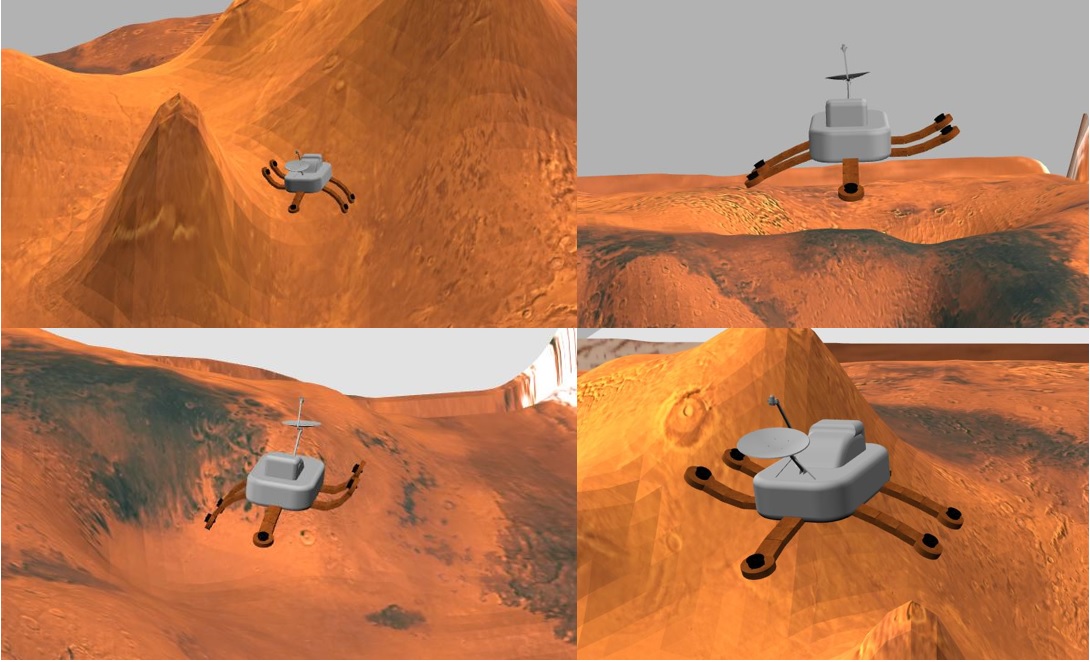}
\caption{Gazebo Simulations of the Soft Morphing Aerial Robot Tilting Its Deformable Arms for Efficient Movement Without Body Rotation, Covering Long Distances in Titan's Sotra Patera Region. }
\label{figure1}
\vspace{-4mm}
\end{figure}

When the drone moves forward, the induced power is:

\begin{equation}
P_{\text{I}} = P_{\text{I\_h}} / \sqrt{1 + (v/(2\sqrt{2}\sqrt{(m g)/(\rho A)}))^2}
\end{equation}

Where:
\begin{itemize}
    \item \(m\) is the drone's mass.
    \item \(g\) is the acceleration due to gravity.
    \item \(A\) is the rotor disk area.
\end{itemize}

\textbf{Profile Power $P_{\text{P}}$ :} This is associated with the aerodynamic drag of the rotating rotor blades. Its expression is:

\begin{equation}
P_{\text{P}} = \rho v^3  A C_Db ( \sigma / 8)
\end{equation}

Where:
\begin{itemize}
    \item $C_Db$ is the drag coefficient of the rotor blades.
    \item \(\sigma\) (solidity) is the ratio of the total blade area to the rotor disk area.
\end{itemize}

The overall power requirement can be aggregated as:

\begin{equation}
P_{\text{total}} = P_{\text{D}} + P_{\text{I}} + P_{\text{P}}
\end{equation}

In Figure 10, the power consumption results for flight on Titan with deformable arms are presented, taking into account the corresponding reduction in drag coefficient and frontal area (see Table II). It can be observed that the total power consumption has a minimum around 13.9 m/s, which is up to 22\% higher than in the case of non-deformable arms without a fully-actuated maneuver. The total consumed power at this optimal point, for the deformable arms scenario, is up to 28\% lower, not exceeding 1.7 kW. 

Finally, considering the battery selected later in Section V.D, with a capacity of 14 kWh, the operational range at optimal velocities increases from 55 km in traditional flight to up to 74 km in the fully-actuated mode. This is of significant importance for enhancing the exploration capabilities of the robot on Titan.

\begin{figure}
\includegraphics[width=0.492\textwidth,scale=0.25]{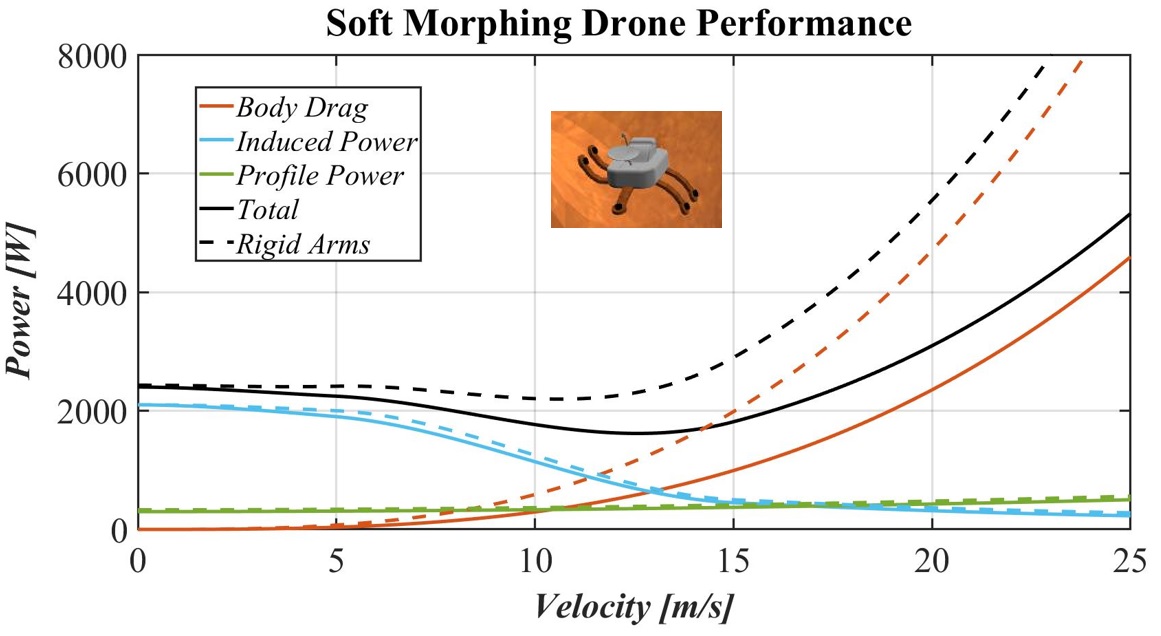}
\caption{Energy consumption assessment of the soft morphing aerial robot in Titan's atmosphere, differentiating powers due to body drag, profile, and induced effects. The total power showcases a minimum for the optimal horizontal travel speed.}
\label{figure1}
\vspace{-4mm}
\end{figure}

\begin{table}[h]
\centering
\caption{Comparison of Flight Characteristics for the rigid and deformable arms in fully-actuated configuration}
\begin{tabular}{|c|c|c|}
\hline
 Parameters & Rigid Arms & Deformable Arms \\
\hline
Optimal Velocity (m/s) & 10.8 & 13.9 \\
\hline
Power (kW) & 2.3 & 1.67 \\
\hline
Drag Coefficient $C_D$ & 1.32 & 0.93 \\ % Replace X and Y with the actual values
\hline
Operational Range (km) & 55 & 74 \\
\hline
\end{tabular}
\end{table}

\vspace{-4mm}

\subsection{Power budget}

In Titan's challenging environment, solar power is significantly hindered, with sunlight being more than 100 times weaker than on Earth. However, Titan's cold, dense atmosphere, which is at a brisk 94-K, requires continuous heat for equipment management. Instruments on the proposed exploration vehicle rely on the residual heat from a Multi-Mission Radioisotope Thermoelectric Generator (MMRTG) for their operations. This heat keeps crucial systems, like the battery, at optimal temperatures.

The expected output of this MMRTG on Titan \cite{Dragonfly} would be around 75W, comparable to the power used by past Mars landers. Consequently, the maximum battery size would be one that fully harnesses the MMRTG energy throughout Titan's nighttime (192 hours), amounting to 75 multiplied by 192, which is 14 kWh; assuming a representative specific energy metric for space-qualified batteries of 100 Wh/kg. For this application, compared to the Dragonfly mission, a 25 kg battery is chosen, which is slightly lighter.

\begin{table}[t]
\caption{Power budget for the soft aerial robot for Titan exploration}
\centering
\begin{tabular}{c c c}
\hline
\textbf{Activity} & \textbf{Power (W)} & \textbf{Duration (h/year)} \\
\hline
Atmospheric Flight  & 1.6 kW & 60 \\
Communications  & 350 & 300 \\
Sample Acquisition & 1.2 kW & 12 \\
Chemical Analysis & 800 & 25 \\
Other Experiments & 200 & 100 \\
Residual Consumption & 30 & Continuous \\
\hline
\end{tabular}

\vspace{-4mm}
\end{table}

The soft aerial robot explorer is tasked with diverse operations (see Table III), each with distinct power demands. Notably, the atmospheric flights stand out, drawing a robust 1.6 kW for 60 hours each year. However, activities like data transmission, sample acquisition, and chemical analysis have varying but crucial energy requirements, all detailed in the table provided. With the MMRTG offering a continuous 100 W and supplemented by the 25 kg battery storing 2.5 kWh, the mission possesses ample energy reserves. The reduction of 28 $\%$ in consumption during forward flight allows the requirements to be significantly reduced.

%\begin{figure}
%\includegraphics[width=0.492\textwidth,scale=0.25]{Figures/Figure_10_iac.jpg}
%\caption{A novel concept for Titan robotic exploration based on soft morphing aerial robots }
%\label{figure1}
%%\end{figure}

\vspace{-1mm}
\section{CONCLUSIONS}

This work introduces a novel concept for Titan robotic exploration through a rotorcraft planetary lander which employs adaptable flexible materials. By controlling the deformation of the multirotor arms, this robotic explorer can achieve full-body perching, allowing it to settle on uneven terrains, thus unlocking new exploration horizons and sampling possibilities. Moreover, as the rotors are set within the deformable material, the interaction with the environment is smoother. The robot's folding feature also eases the demands on hypersonic aeroshell designs.

Understanding the behavior of flexible materials in Titan's extreme environment emerged as one of the primary challenges of this concept. As a result, this study involved testing small-scale deformable arms using liquid nitrogen to monitor their stability under cryogenic temperatures. It was found that PTFE, despite having a higher modulus of elasticity than TPU or silicone, maintains its characteristics across a broad temperature spectrum. The recent advancements in 3D printing of such materials, despite their high melting points, make them particularly appealing.

Additionally, the ideal rotor configuration, using a Ducted Fan approach suitable for Titan's atmospheric conditions, has been determined, yielding promising efficiency outcomes. This, combined with Gazebo simulations of the soft morphing aerial robot's behavior, supports the proposed design. Forward flight energy consumption has been reduced by more than 25 $\%$ due to fully-actuated efficient flight with morphing arms, allowing the longer-range explorer robot to reach unexplored areas more easily.

The in-depth conceptual design of this aerial robot holds significant potential for Titan exploration, specifically in the region of the Sotra-Patera cryovolcano. Such exploration could provide insights into Titan's methane cycle and its similarities to Earth. In conclusion, this concept enhances overall safety and exploration potential, making it promising for future exploration of Titan.

\section*{ACKNOWLEDGMENTS}
%(Aerial Robotic Training for the next generation of European infrastructure and asset maintenance technologies)- European Training Network (ETN) project 
We thank Robotics, Vision and Control Group (GRVC). This work has been developed within the framework of the AERO-TRAIN project, a Marie-Sklodowska-Curie Innovative Training Network (ITN)  (Grant agreement 953454).

%%%%%%%%%%%%%%%%%%%%%%%%%%%%%%%%%%%%%%%%%%%%%%%%%%%%%%%%%%%%%%%%%%%%%%%%%%%%%%%%

\vspace{-3mm}

\bibliographystyle{IEEEtran}
\bibliography{References}

\end{document}